\documentclass[sigchi]{acmart}

\usepackage{booktabs} % For formal tables

\usepackage{balance}
\AtBeginDocument{%
  \providecommand\BibTeX{{%
    \normalfont B\kern-0.5em{\scshape i\kern-0.25em b}\kern-0.8em\TeX}}}

\setcopyright{acmcopyright}
\copyrightyear{2019}
\acmYear{2019}
\acmDOI{10.1145/3340555.3355716}
\acmConference[ICMI '19]{International Conference on Multimodal Interaction}{October 14--18, 2019}{Suzhou, China}
\acmBooktitle{2019 International Conference on Multimodal Interaction (ICMI '19), October 14--18, 2019, Suzhou, China}
\acmPrice{15.00}
\acmISBN{978-1-4503-6860-5/19/10}

\begin{document}

    \title{Automatic Group Cohesiveness Detection With Multi-modal Features}

\author{Bin Zhu}
\orcid{1234-5678-9012}
\affiliation{%
  \institution{Department of Electrical and Computer Engineering, University of Delaware}
  \city{Newark}
  \state{DE} 
  \postcode{19716}
  \country{USA}
  }
\email{zhubin@udel.edu}

\author{Xin Guo}

\affiliation{%
  \institution{Department of Electrical and Computer Engineering, University of Delaware}
  \city{Newark} 
  \state{DE} 
  \postcode{19716}
  \country{USA}
}
\email{guoxin@udel.edu}

\author{Kenneth E. Barner}
\affiliation{%
  \institution{Department of Electrical and Computer Engineering, University of Delaware}
  \city{Newark}
  \state{DE} 
  \postcode{19716}
  \country{USA}
  }
\email{barner@udel.edu}

\author{Charles Boncelet}
\affiliation{%
  \institution{Department of Electrical and Computer Engineering, University of Delaware}
  \city{Newark}
  \state{DE} 
  \postcode{19716}
  \country{USA}
  }
\email{boncelet@udel.edu}

% \author{Bin Zhu \qquad Xin Guo \qquad Kenneth E. Barner\qquad Charles Boncelet}
% \affiliation{%
%  \institution{Department of Electrical and Computer Engineering, University of Delaware}
%  \city{Newark}
%  \state{DE} 
%  \postcode{19716}
%  \country{USA}
%  }
% \email{{zhubin, guoxin, barner, boncelet}@udel.edu}
%
% By default, the full list of authors will be used in the page
% headers. Often, this list is too long, and will overlap
% other information printed in the page headers. This command allows
% the author to define a more concise list
% of authors' names for this purpose.
\renewcommand{\shortauthors}{Zhu and Xin, et al.}

\begin{abstract}
 Group cohesiveness is a compelling and often studied composition in group dynamics and group performance. The enormous number of web images of groups of people can be used to develop an effective method to detect group cohesiveness. 
This paper introduces an automatic group cohesiveness prediction method for the 7th Emotion Recognition in the Wild (EmotiW 2019) Grand Challenge in the category of Group-based Cohesion Prediction. The task is to predict the cohesive level for a group of people in images. To tackle this problem, a hybrid network including regression models which are separately trained on face features, skeleton features, and scene features is proposed. Predicted regression values, corresponding to each feature, are fused for the final cohesive intensity. Experimental results demonstrate that the proposed hybrid network is effective and makes promising improvements. A mean squared error (MSE) of 0.444 is achieved on the testing sets which outperforms the baseline MSE of 0.5.
\end{abstract}

\begin{CCSXML}
<ccs2012>
<concept>
<concept_id>10010147.10010178.10010224.10010225.10010228</concept_id>
<concept_desc>Computing methodologies~Activity recognition and understanding</concept_desc>
<concept_significance>500</concept_significance>
</concept>
<concept>
<concept_id>10010147.10010178.10010224.10010225.10010227</concept_id>
<concept_desc>Computing methodologies~Scene understanding</concept_desc>
<concept_significance>300</concept_significance>
</concept>
<concept>
<concept_id>10010147.10010257.10010258.10010262.10010277</concept_id>
<concept_desc>Computing methodologies~Transfer learning</concept_desc>
<concept_significance>300</concept_significance>
</concept>
</ccs2012>
\end{CCSXML}

\ccsdesc[500]{Computing methodologies~Activity recognition and understanding}
\ccsdesc[300]{Computing methodologies~Scene understanding}
\ccsdesc[300]{Computing methodologies~Transfer learning}

\keywords{Cohesion Prediction, Multi-modal features, Transfer learning }

\maketitle
\section{Introduction}
Group Cohesiveness plays an important role in the study of small group behavior, social psychology, group dynamics, sport psychology, and organizational behavior \cite{doi:10.1177/104649648001100401,doi:10.1080/14792779343000031}. Cohesiveness has been found to be one of the critical influencing factors in group performance. Several studies have shown that strong group performance is associated with a high level of group cohesion among the members \cite{articleBanwo,doi:10.1080/09718923.2005.11892479}. Moreover, recent research \cite{Ghosh2019Cohesiveness} shows that group cohesion is highly correlated to group-level emotion.

The rapid growth of web images, driven by photo hosting and sharing services such as Flickr, FaceBook, and Google Photos, has gradually and significantly changed our life style \cite{inproceedingsMiller}. Many of these images are taken when people are attending meaningful social events, such as graduations, birthday parties, and family gatherings. Such images not only capture these most precious moments, but also have useful information that can be used to analyze group-level social attributes such as group cohesion. The availability of these images motivates the design of automatic systems capable of understanding human perception of cohesion at the group level.

Measuring and annotating group cohesion at different levels is often difficult for a human annotator, because cohesion has team and individual components \cite{articleSalas}. The problem of group cohesiveness prediction becomes even more challenging in static images.  Complications include face occlusions, illumination variations, head pose variations, varied indoor and outdoor settings, faces at different distances from the camera, and low-resolution face images. In this paper, we propose a robust ensemble model that separately processes various high-level information of faces, skeletons, and scenes. Then, regression values are calculated and fused for the final cohesive intensity. In the 7th Emotion Recognition in the Wild (EmotiW 2019) Sub-Challenge \cite{Dhall:2019}, the proposed hybrid model achieves a competitive result.

\begin{figure*}
\includegraphics[height=6.8cm,width=0.90\textwidth]{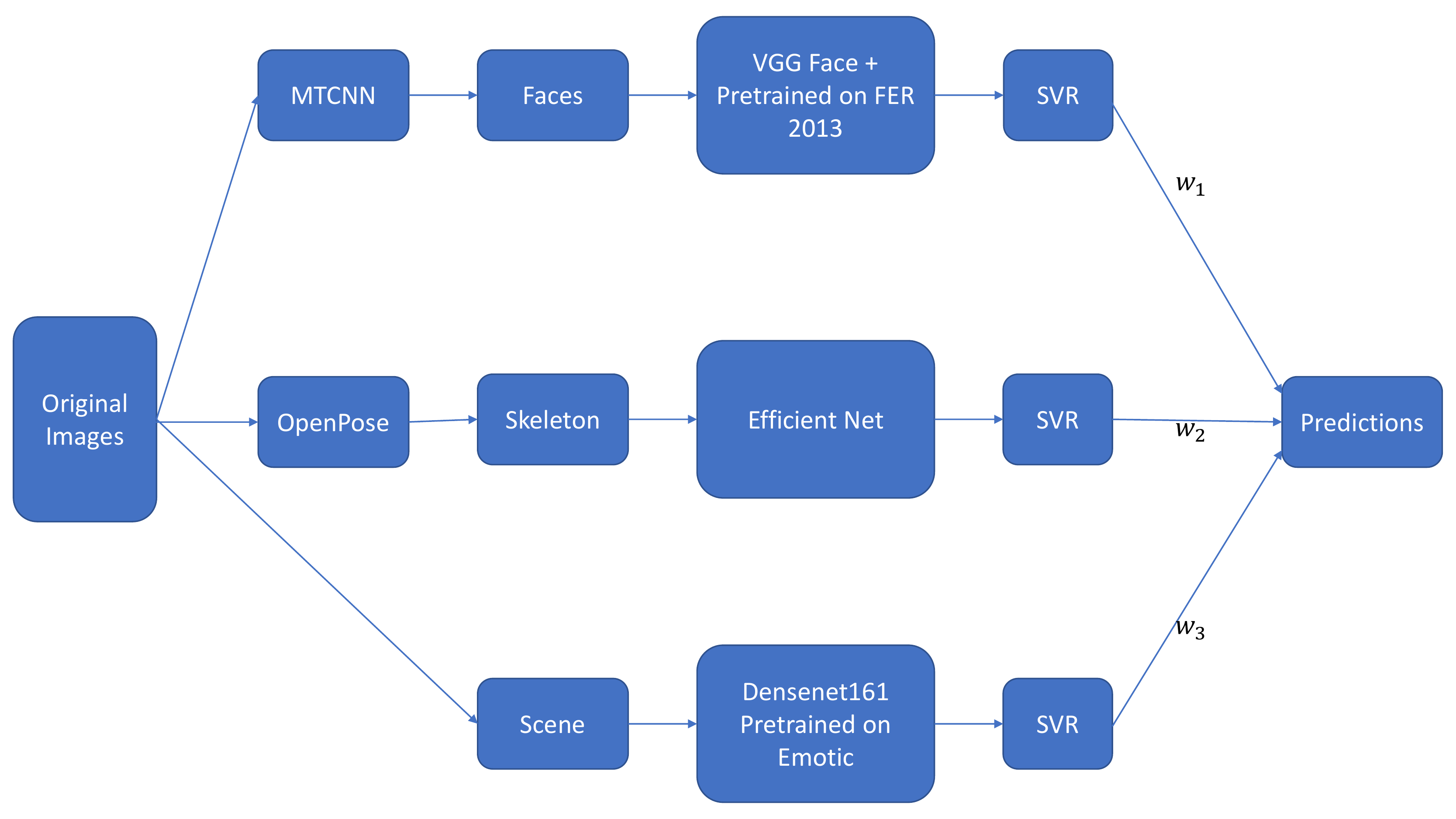}
\caption{Overall Proposed Hybrid Network structure.}
\label{fig2}
\end{figure*}

\section{Related Work}

Many researchers have employed the rapidly developing computer vision and machine learning techniques to machine understanding of images and videos.  One specific task is to study groups of people from images.

Photos of groups of people during social gatherings, such as birthday parties, graduations, and family reunions, are widely available.  \cite{5206828} introduces contextual features that capture the structure of a group of people and the position of individuals within the group. This social context helps to accomplish a variety of tasks such as the following: identifying the demographics of people in the group, estimating camera and scene parameters, and classifying group events. 

% A video tracking system based on group-level activity recognition in complex and crowded environments such as schools, prisons, and public parks is presented by \cite{5597317}. To tackle the problem, hierarchical agglomerative clustering and hierarchical divisive clustering using modularity cut are implemented. These methods are integrated in a comprehensive real-time surveillance system that performs multi-camera and multi-target tracking, effectively recognizing both low-level and high-level activities such as group loitering and group agitation respectively. 

Recently, the EmotiW 2019 Challenge organizers presented the first study of group cohesion prediction in static images \cite{Ghosh2019Cohesiveness}. The challenge organizer extends the  Group  Affect  Database  \cite{Dhall:2017:IGE:3136755.3143004} with  group  cohesion  labels  and  proposes the new GAF Cohesion database. Two deep cohesion models, separately trained on holistic and face-level features, achieve results on the Cohesion database which approximate human-level performance. Motivated by considering cohesiveness as an attribute of group emotion, the paper jointly trains an inception V3 model on both group emotion and group cohesion. From the experimental results, joint training on both emotion and cohesion achieves a higher performance than individual training. It strongly infers that group emotion and cohesion are correlated.

\section{Proposed Method}
The system pipeline is shown in Figure 1. The basic idea of the proposed approach is to train a Support Vector Regression (SVR) \cite{articleVapnik} with high-level features of the input images from different representations. The predicted regression values are fused by using a grid search to achieve the final prediction.

\subsection{Scene Features}
Holistic (scene-level) information is shown to be the important component in group-level classification in \cite{7298768,Guo:2018:GER:3242969.3264990,Guo17}. While analyzing the cohesiveness of a group of people, it is essential to understand the environments behind the people, e.g., students in a lecture tend to have a low cohesion level, while a group people standing and protesting at a plaza probably have high cohesiveness.
In order to extract the high-level interpretations of the holistic information, a state of art deep model Densely Connected Convolutional Network (DenseNets) \cite{DBLP:journals/corr/HuangLW16a} is applied.

DenseNets have several important advantages: alleviating the vanishing-gradient problem, strengthening feature propagation, feature reusing, and substantially reducing the number of parameters. DenseNets accomplish significant improvements over the state-of-the-art on four highly competitive object recognition benchmark tasks (CIFAR-10, CIFAR-100, SVHN, and ImageNet).
Moreover, before extracting holistic features by using DenseNets, we fine-tune the Densenents network on Emotic Dataset \cite{emotic_cvpr2017}. Group cohesion level is relevant to the group-level emotion or valance degree. The Emotic Dataset consists of a total of 18,316 images that are labeled in two methods, 26 emotion discrete categories, and valence continuous dimensions scaled from 1 to 10. A pretrained (on Imagenet) DenseNet161 model is fine-tuned by using the Emotic Dataset labeled in continuous dimensions. With the exception of the last layer, a size 2208 feature vector is extracted for each original image.

\subsection{Face Feature}
Considering the high correlation between group-emotion and group cohesion, the overall facial emotion stage of a group of people can contribute to group cohesiveness detection. The sample images shown in Figure 2 demonstrate that the average facial expression among all faces is a substantial indicator of group cohesiveness in the image. For instance, if most of the faces are classified as neutral expressions, the group cohesion level tends to have a lower value. In such a manner, faces are extracted by using Multi-task Cascaded Convolutional Network (MTCNN) which is effectively detecting and aligning faces in real time and achieves superior accuracy on the challenges FDDB and WIDER FACE benchmark for face detection and AFLW benchmark for face alignment \cite{Zhang_2016}. 

The VGG Face is a deep network, containing 22 layers and 37 deep units, trained on a very large scale dataset \cite{Parkhi2015DeepFR}. This dataset contains 2.6M images with over 2.6K people which is assembled by a combination of automation and manual operations. The fine-tuned VGG Face model is often used as a feature extractor to extract the activation vector of the fully connected layer in the CNN architecture. It has proven more efficient than a trained from scratch model \cite{Guo_2016,8373900}. In furtherance of exploiting the high-level abstractions of extracted faces, the VGG Face model is trained on the facial expression dataset FER 2013 \cite{DBLP:journals/fer2013}. Then, VGG Face considered as a feature extractor with the last fully connected layer removed, computes a size of 4096 feature vector for all faces. Moreover, we obtain a different representation for each face. To train our SVR model, a single representation of each image is required. However, simply concatenating all feature vectors is invalid because each image can consist of a different number of faces. In this way, the face feature vectors are averaged to obtain a single facial feature vector to feed into the SVR predictor.

\subsection{Skeleton Feature}
 As shown in Figure~\ref{fig3}, skeleton features demonstrate salient patterns of different categories through facial expressions, poses, gestures, and the structures of groups of people. In this work, the skeleton of each image is extracted using OpenPose~\cite{cao2017realtime, simon2017hand,wei2016cpm}, which can jointly detect human body, hand, and facial keypoints (in total 135 keypoints) on each image. Furthermore, the Openface library contains multiple functions such as 2D real-time multi-person keypoint detection, 3D real-time single-person keypoint detection, a calibration toolbox, and single-person tracking.

A new model, EfficientNet, achieves state-of-the-art accuracy on ImageNet, CIFAR-100, and Flowers, while being 8.4x smaller and 6.1x faster on inference than the best existing ConvNet \cite{pmlr-v97-tan19a}. EfficientNet is powered by a novel scaling method and the  advanced Automated machine learning (AutoML). The heuristic model scaling method uses a simple yet highly effective compound coefficient to scale up CNNs in a more structured manner. Moreover, this method uniformly scales each dimension with fixed scaling coefficients. This scaling is different from traditional approaches, e.g., ResNet arbitrarily scales up layers from Resnet-18 to Resnet-50, Resnet-101 and Resnet-152, while they usually require tedious manual tuning. A pre-trained (on Imagenet) EfficentNet model, with the exception of the last layer, extracts a size of 1536 feature vector for each original image.

\begin{figure}
\includegraphics[height=5.5cm, width=1.\columnwidth]{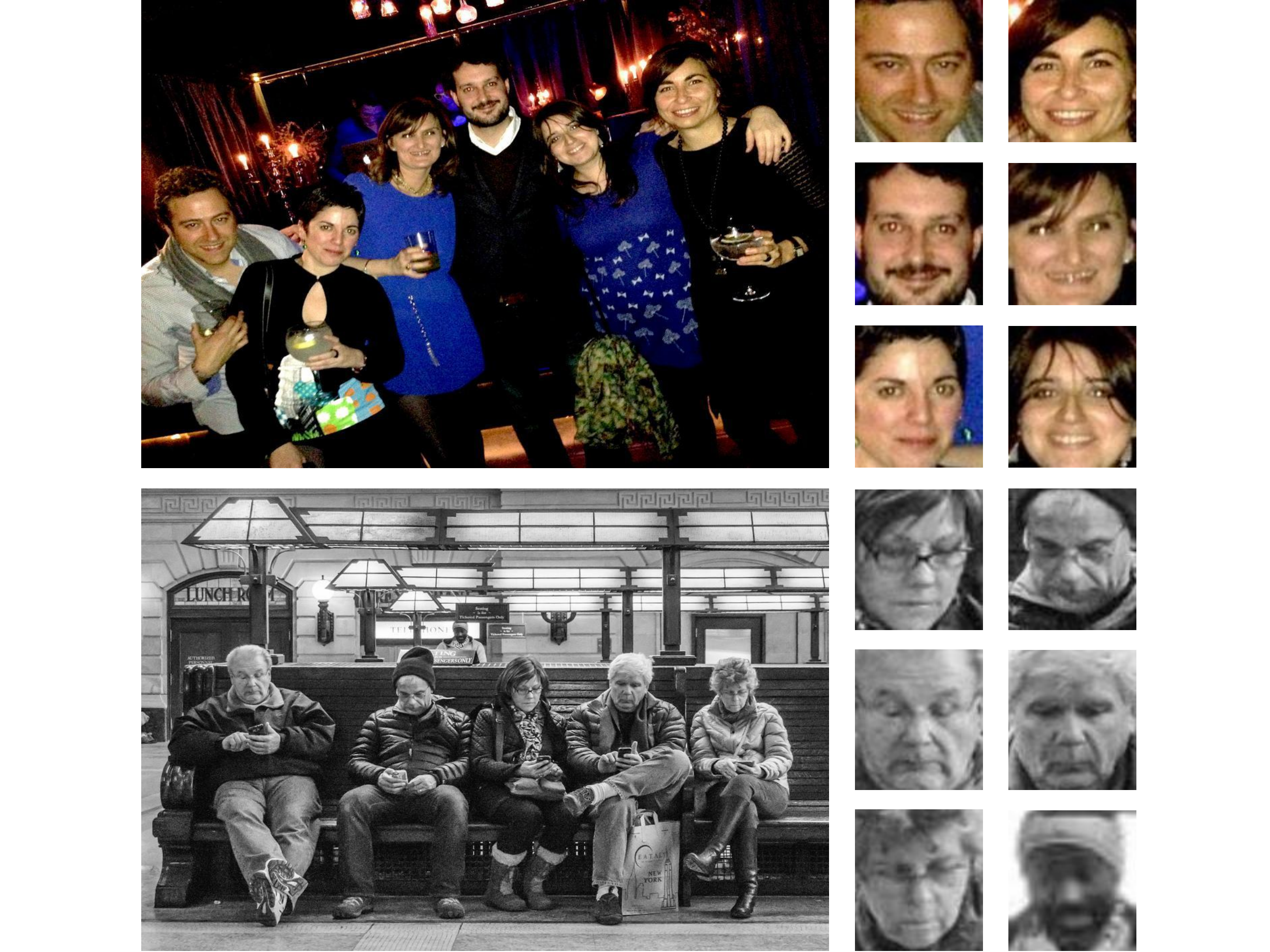}
\caption{Samples of faces. Top: High-level Group Cohesiveness Below: Low-level Group Cohesiveness. }
\label{fig2}
\end{figure}

\begin{figure}
\includegraphics[width=0.9\columnwidth]{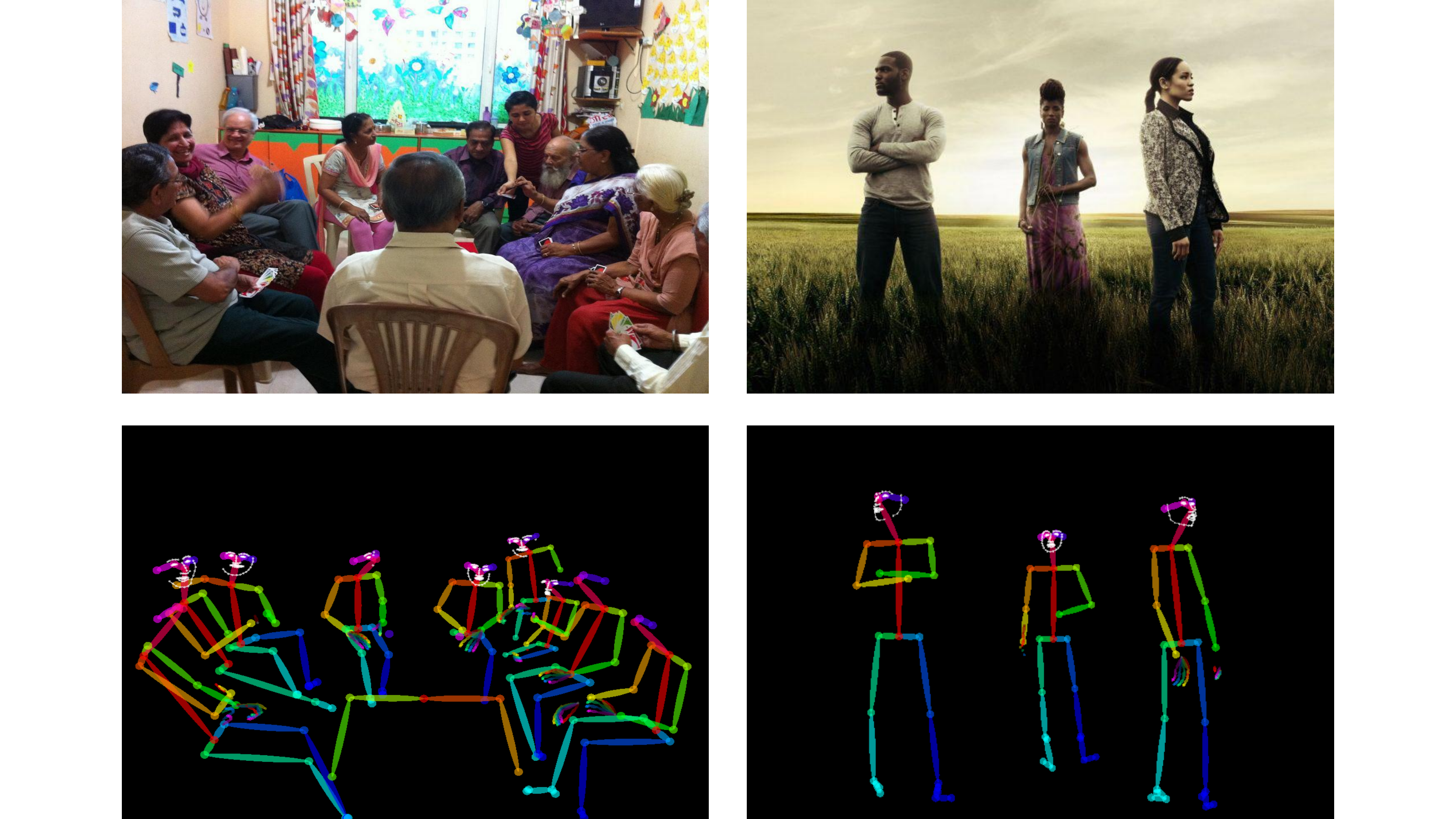}
\caption{Samples of skeleton feature representations. Left: High-level Group Cohesiveness Right: Low-level Group Cohesiveness. }
\label{fig3}
\end{figure}

\section{Experiments}
\subsection{Dataset}

The group cohesiveness prediction dataset in Emotiw 2019 contains a total of 14,175 images. It is split into three parts: 9815 images for training, 4,349 images for validation, and 3011 images for testing. The database consists of all images in GAF 3.0 database \cite{Dhall:2017:IGE:3136755.3143004}, and new set of images are added and collected via web crawlers with various keywords related to social activities, e.g., wedding, birthday party, riot, and protest, etc. The dataset is labeled in four categories as cohesive level 0, 1, 2 and 3.

To better understand the perception of group cohesion and improve the labeling of the dataset, the Emotiw 2019 Challenge conducted a survey via a Google form with 102 participants (59 male and 43 female) whose age ranges from 22 to 54. The survey contained 24 images of groups of people in different contexts and has 4 different Group Cohesion Score (GCS) values. The participants selected one of GCS values for each image and described reasons behind their choice by using provided keywords related to the AGC score.

With the assistance of the survey results, we employed 5 annotators (3 females and 2 males) labeling each image for its cohesiveness in the range [0,3].

\subsection{Experiment setting}
The deep networks (DenseNet, EfficientNet and VGG FACE) are implemented in Pytorch powered by NVIDIA GFORCE 1080. The original images are resized to 224x224 to fit the CNNs as input, and the provided labels are normalized from [0, 3] to [0, 1]. After reviewing the training dataset, we notice that the dataset is severely imbalanced. The distribution of the training dataset is as follows: 1141 images belong to level 0, 1561 for level 1, 4601 for level 2, and 1997 for level 3. To balance the data, 30\%   of the images from the category of level 2 are down-sampled. 

\begin{table}
  \caption{Performance on the validation set.}
  \begin{tabular}{ccc}
    \toprule
    Method & Dataset  & MSE \\
    \midrule
    Baseline & Train  & 0.84\\
    Face& Train & 0.703  \\
    Skeleton& Train& 0.775 \\
    Scene& Train&  0.731\\
    Fusion+Average & Train  & 0.691\\
    Fusion+Grid Search &  Train & 0.683\\
    Face& Balanced Train & 0.678\\
    Fusion+Average & Balanced Train & 0.672\\
    Fusion+Grid Search&Balanced Train & 0.662\\
  \bottomrule
\end{tabular}
\label{tab1}
\end{table}

\subsection{Experiment results}
We conduct experiments on both original training set and balanced training set, and the table 1 shows the validation results. As shown in table 1, our fusion model significantly decreases the MSE. Due to the bias in the training data, data augmentation is important in this challenge and we achieve the lowest MSE of 0.662 on validation set by using our proposed approach with balanced training data.
For the test phase, we use the fusion model which achieves the best result on validation. Table 2 summarizes our 5 submission results. Table 3 presents submission results of MSE corresponding to each individual cohesive level. To make use of all available data, we combine both training data and validation data to train our model. However, the performance is decreased, and submission 2 and submission 5 demonstrate the conclusion. The possible reason is the combined data without modification are severely biased which causes model over-fitting.  Eventually, in submission 4, our model achieves the best MSE 0.444 on combined data with data augmentation.

\begin{table}
  \caption{Submission Results}
  \begin{tabular}{cccc}
    \toprule
   Sub& Method &Dataset&  Test MSE\\
    \midrule
    1 & Fusion+Average & Train&  0.466\\
 
    2 & Fusion+Average& Train + Val&  0.478\\
 
    3 & Fusion+Average & Balanced Train + Val &0.466\\
 
    4 & Fusion+Grid Search & Balanced Train + Val  & 0.444\\
 
    5 & Fusion+Grid Search & Train + Val & 0.447\\
      \bottomrule   
        
  \hline
\end{tabular}
\label{tab2}
\end{table}

% \begin{table}
%   \caption{Submission Results(MSE on Individual Level) }
%   \begin{tabular}{ccccc}
%     \toprule
%   Sub& level 0 & level 1& level 2 &level3\\
%     \midrule
%     1 & 2.86&0.733&0.0689&0.781\\
 
%     2 & 2.52&0.606&0.0952&0.853\\
 
%     3 & 2.81&0.706&0.0823&0.771\\
 
%     4 & 2.28&0.632&0.144&0.684\\
 
%     5 & 2.30&0.616&0.130&0.717\\
%       \bottomrule   
        
%   \hline
% \end{tabular}
% \label{tab3}
% \end{table}

% \section{Discussion}
% To improve the proposed method, multiple innovative techniques are also investigated and implemented. An attention mechanism is applied to fuse the facial features, because the importance of faces in an image is different, e.g. some faces are captured entirely while some are not. Several studies have shown an attention mechanism achieves compelling performance on many different types of tasks \cite{Genta18,Ashish17,Aarush18}. Another novel data augmentation technique is applied and examined in our approach to overcome the imbalanced data and over-fitting problem.  The Random Erasing data augmentation method has shown to be an effective complement to commonly used data augmentation methods such as random cropping and flipping \cite{DBLP:journals/corr/abs-1708-04896}. However, these two methods we investigated did not improve our proposed method. One possible reason is that most extracted faces have very low resolution. Although the Random Erasing technique can improve robustness of the model, determinant information in the image may also erased. 
\section{Conclusion}

In summary, group cohesiveness is a major component for analyzing group behavior, group performance, group emotion etc. A large number of images, taken from social gathering and social activities, are shared on online photo services such as Flickr and Facebook.

In addition, measuring and annotating group cohesion at different levels for a human annotator is usually time consuming and inefficient. In this paper, we construct a robust ensemble hybrid regression model to automatically and effectively detect group cohesiveness. The model is separately trained on faces, skeletons, and scenes. The regression values are fused for the final cohesive intensity. Our experiments deliver a mean squared error of 0.662 and 0.444 on the validation and testing sets, respectively. This MSE outperforms the baseline MSE of 0.5. The result demonstrates that the proposed hybrid model is effective and makes promising improvements.


\bibliographystyle{ACM-Reference-Format}
\bibliography{cohesion}


\end{document}